\documentclass{article} 
\usepackage{iclr2026_conference,times}

\makeatletter
\let\AddToShipoutPicture\relax 
\makeatother

\usepackage[hang,flushmargin]{footmisc}
\usepackage{fancyhdr}
\fancypagestyle{plain}{\fancyhf{}} 
\usepackage{amsmath,amsfonts,bm}

\def\eqref#1{equation~\ref{#1}}

\def\1{\bm{1}}

\DeclareMathAlphabet{\mathsfit}{\encodingdefault}{\sfdefault}{m}{sl}
\SetMathAlphabet{\mathsfit}{bold}{\encodingdefault}{\sfdefault}{bx}{n}

\usepackage{hyperref}
\usepackage{url}

\usepackage{wrapfig}   
\usepackage[table,xcdraw]{xcolor}  
\usepackage{graphicx}  
\usepackage{booktabs}  
\usepackage{tikz}
\usepackage{pifont}
\usepackage{amsmath}
\usepackage{enumitem}
\usepackage{longtable}
\usepackage{microtype}
\usepackage{subfig}
\usepackage[section]{placeins}
\usepackage{comment}
\usepackage{multirow}
\usepackage{caption}
\usepackage{relsize}
\usepackage{colortbl} 
\usepackage{arydshln} 
\usepackage{algorithm}
\usepackage{tcolorbox} 
\usepackage{wrapfig}
\usepackage{subcaption}
\usepackage{diagbox}
\usepackage{makecell}
\usepackage{collcell}
\usepackage{pgf}
\usepackage[dvipsnames]{xcolor}
\usepackage{textcomp}
\usepackage{amssymb}
\usepackage{mathtools}
\usepackage{siunitx}
\definecolor{lightgreen}{RGB}{0,150,0}  
\definecolor{lightblue}{RGB}{173,216,230}  
\newlength{\Dcolw}
\title{EvoSyn: Generalizable Evolutionary Data Synthesis for Verifiable Learning}

\author{
\textbf{He Du}\textsuperscript{\rm 1 \rm 2}\quad
\textbf{Bowen Li}\textsuperscript{\rm 2}\footnotemark[1]\quad
\textbf{Aijun Yang}\textsuperscript{\rm 2 }\quad
\textbf{Siyang He}\textsuperscript{\rm 2}\quad
\textbf{Qipeng Guo}\textsuperscript{\rm 2}\thanks{Corresponding author.}\quad
\textbf{Dacheng Tao}\textsuperscript{\rm 3}\vspace{0.1cm}
\\
\textsuperscript{\rm 1}Fudan University \quad
\textsuperscript{\rm 2}Shanghai AI Laboratory \quad 
\textsuperscript{\rm 3}Nanyang Technological University \vspace{0.1cm}
\\
~~\texttt{elyndendu@gmail.com}~,~\texttt{libowen.ne@gmail.com}
}

\iclrfinalcopy 
\begin{document}

\maketitle

\begin{abstract}
   Reliable verifiable data has become a key driver of capability gains in modern language models, enabling stable reinforcement learning with verifiable rewards and effective distillation that transfers competence across math, coding, and agentic tasks. Yet constructing generalizable synthetic verifiable data remains difficult due to hallucination-prone generation, and weak or trivial verification artifacts that fail to separate strong from weak solutions. Existing approaches often rely on task-specific heuristics or post-hoc filters that do not transfer across domains and lack a principled, universal evaluator of verifiability. In this work, we introduce an evolutionary, task-agnostic, strategy-guided, executably-checkable data synthesis framework that, from minimal seed supervision, jointly synthesizes problems, diverse candidate solutions, and verification artifacts, and iteratively discovers strategies via a consistency-based evaluator that enforces agreement between human-annotated and strategy-induced checks. This pipeline upgrades filtering into principled synthesis: it reliably assembles coherent, verifiable training instances and generalizes without domain-specific rules.
   Our experiments demonstrate the effectiveness of the proposed approach under both RLVR and model distillation training paradigms. The results show that training with our synthesized data yields significant improvements on both the LiveCodeBench and AgentBench-OS tasks, highlighting the robust generalization of our framework. The code and data are available at~~\url{https://github.com/kinza99/openevolve}.
\end{abstract}

\section{Introduction}
Large language models (LLMs) have demonstrated remarkable potential across a wide range of domains, particularly in complex reasoning tasks such as mathematics, programming, and real-world agent applications. 
Recently, models like OpenAI-o1 and DeepSeek-R1~\citep{guo2025deepseek,gpt_o1,yang2025qwen3}, after undergoing large-scale reinforcement learning, have shown significant improvements on reasoning benchmarks~\citep{yue2025doesreinforcementlearningreally,su2025crossingrewardbridgeexpanding}.
However, as model capabilities rapidly advance, their size continues to grow, and their demand for data is expanding at an astonishing pace. 
In particular, recent training paradigms increasingly rely on a special class of data—verifiable data.

Verifiable data provides reliable feedback signals during training, making it indispensable for many approaches. For example, RLVR-style training methods and model distillation heavily rely on such data~\citep{schulman2017proximal,shao2024deepseekmath,zhao2025absolute}; DPO~\citep{hosseini2024v,lai2024step} leverages feedback to construct positive and negative samples; and various self-training methods such as STaR~\citep{zelikman2022star}, V-STaR~\citep{hosseini2024v}, and ReST~\citep{rest_em} all depend on correctness signals to filter useful examples. However, the stringent reliability requirements of verifiable data make it extremely costly to annotate. Large-scale manual labeling is simply infeasible, highlighting the growing importance of verifiable data in modern LLM training pipelines.

Synthetic data offers a promising solution, but it remains imperfect~\citep{liu2024best,long2024llms,naduaș2025synthetic}. Two persistent challenges limit its utility.
First, \textit{reliability}: hallucinations remain a fundamental weakness of LLMs. While models can generate large volumes of data, ensuring their reliability is nontrivial~\citep{ding2024dataaugmentationusinglarge, he-etal-2025-fine}.
How to make model-generated data more reliable or how to effectively filter trustworthy subsets from large synthetic corpora remains a central challenge.
Second, \textit{generalizability}: Many existing solutions rely on task-specific, handcrafted heuristics to guarantee data usability. For example, some studies validate correctness through syntax checking~\citep{wang2025codecontests+}. These approaches, however, often fail to generalize beyond the narrow task domains they were designed for.

In this work, we focus on these two questions: how to obtain reliable, verifiable data, and how to design a unified pipeline that generalizes across diverse tasks. We target the executably-checkable data class, which is the major part of verifiable data. We propose a general-purpose framework for synthesizing reliable data, called Evolutionary Data Synthesis (EvoSyn). Executably-checkable tasks are a broad class of problems defined as those for which verification can be performed via tests without requiring a complete solution. This class encompasses challenging real-world tasks, such as coding and software engineering problems. In our experiments, we select representative and high-difficulty tasks: the algorithmic LiveCodeBench~\citep{jain2024livecodebench} and the complex agent task AgentBench-OS~\citep{liu2023agentbench}. The core idea of EvoSyn is to formulate the difficulty as a data filtering strategy optimization task. Inspired by AlphaEvolve~\citep{novikov2025alphaevolve}, we employ evolutionary algorithms to iteratively search for the optimal filtering strategy tailored to the current task~\citep{openevolve,romera2024mathematical,tanese1989distributed}. This strategy is then applied to synthetic data, yielding a reliable, verifiable dataset.
Unlike prior approaches that require handcrafted, task-specific heuristics, EvoSyn automates this process: the model itself explores and evolves filtering strategies, reducing manual effort while producing superior solutions. Crucially, EvoSyn introduces a unified evaluation criterion for filtering strategies, which is task-agnostic. Instead of relying on domain-specific signals, EvoSyn measures consistency score with a small set of manually verified seed data, making it applicable to any verification task as long as minimal seed supervision is available.

\begin{figure}[h]
   \vspace{-10pt}
   \begin{center}
   \includegraphics[width=0.75\linewidth]{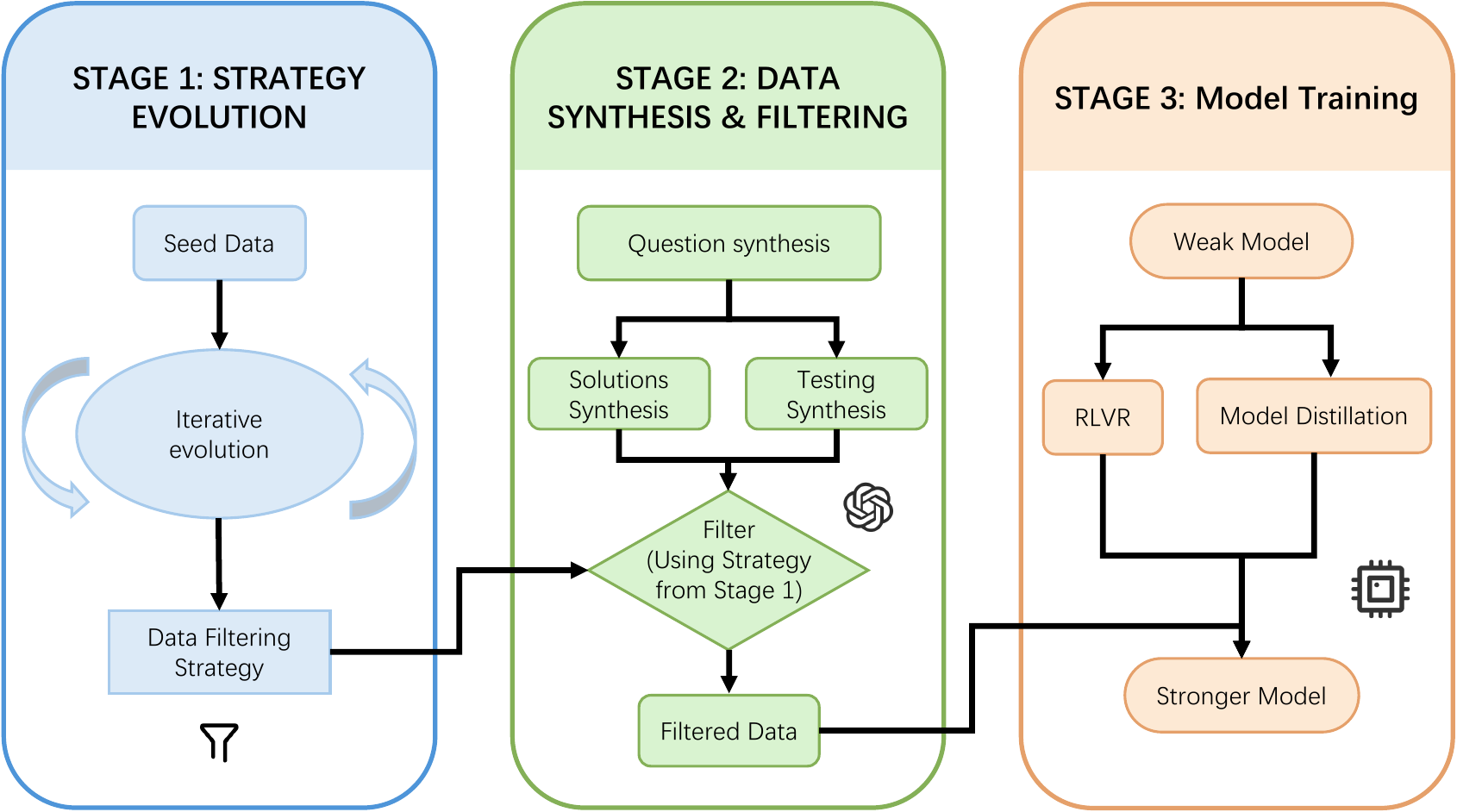}
   \end{center}
   \caption{Overview of EvoSyn, a task-agnostic pipeline for synthesizing verifiable data. From a small human-verified seed data, an evolutionary process discovers a data-filtering strategy via a consistency-based evaluator; this strategy then guides synthesis by generating candidate solutions and tests for new problems, cross-executing them to rank and retain reliable instances while discarding trivial or inconsistent ones. The resulting verifiable dataset (problems, tests, and strong solutions) supports training in diverse tasks.}
   \label{fig:main_fig}
   \vspace{-15pt}
\end{figure}

We demonstrate that EvoSyn is both effective and generalizable. Through its evolutionary process, EvoSyn continuously discovers novel and increasingly powerful strategies over iterations. We showcase representative examples and provide a detailed analysis of how strategy quality improves as the number of evolutionary rounds increases. Next, we validate EvoSyn on model training. 
On LiveCodeBench~\citep{jain2024livecodebench}, we conduct RLVR training, and EvoSyn-generated data significantly improve the performance of LLaMA-3.1~\citep{grattafiori2024llama} and Qwen3~\citep{yang2025qwen3} models, outperforming raw synthetic baselines and providing more effective training dynamics.
On the challenging AgentBench-OS benchmark, we choose the representative model distillation method, EvoSyn also yields substantial gains, enabling distilled models to surpass not only random baselines but also their teacher model (DeepSeek-R1~\citep{guo2025deepseek}).

Our main contributions are:

\text{~~~~~~• }We introduce Evolutionary Data Synthesis (EvoSyn), a general framework for synthesizing verifiable data. EvoSyn automatically evolves a superior data filterering strategies for the given task, enabling the construction of reliable synthetic datasets.

\text{~~~~~~• }We provide a detailed study of EvoSyn's evolutionary process, demonstrating its effectiveness, generalizability, and cost trade-offs.

\text{~~~~~~• }We validate EvoSyn on two important training paradigms, RLVR and model distillation, showing that EvoSyn-generated data yields substantial improvements over baselines.

\section{Related Work}
\label{related_work}

\paragraph{Verifiable learning}
Verifiable learning leverages executable or checkable feedback to supervise model training and spans both RL with verifiable rewards (RLVR)~\citep{lambert2025tulu3pushingfrontiers} and supervised fine-tuning/distillation. In RLVR~\citep{schulman2017proximal,shao2024deepseekmath,guo2025deepseek,gpt_o1,yang2025qwen3}, correctness signals from program execution, unit tests, or other deterministic checkers stabilize training and markedly enhance reasoning ability. Beyond RLVR, teacher outputs can be filtered by execution in model distillation~\citep{kim2025reinforcement}; and self-training pipelines such as RFT, STaR, and ReST~\citep{rest_em,zhang2024rest,zelikman2022star} rely on correctness signals to retain useful data. Verification feedback also constructs preference data for  DPO~\citep{hosseini2024v,lai2024step,rafailov2024direct} and improves reward models~\citep{math-shepherd}.

\paragraph{Data synthesis}
Synthesizing verifiable data is critical yet challenging~\citep{liu2024best,long2024llms,naduaș2025synthetic}. In practice, high-quality data for executably-checkable data often require broad-coverage unit tests~\citep{chen2022codet,wang2025codecontests+}, program-analysis tooling~\citep{liang2025grammar}, or carefully curated exemplars~\citep{shao2024case2code}. Such task-specific heuristics incur high manual costs and transfer poorly to complex real-world reasoning tasks~\citep{fandina2025automated,jimenez2023swe,zhang2025swe,li2024devbench}. Hallucination further undermines reliability, making robust verification artifacts themselves a central bottleneck~\citep{long2024llms}.


\section{Methodology}
\label{methodology}

To address the inherent unreliability of synthetic data, we propose a new approach, \textit{Evolutionary Data Synthesis (EvoSyn)}. EvoSyn targets executably-checkable tasks that satisfy two conditions: (1) correctness can be decided by executable “testing” artifacts (e.g., unit tests, checkers, environment assertions) that deterministically accept or reject candidate solutions; and (2) such testing artifacts can be authored without first producing a correct solution (e.g., via specifications, invariants, metamorphic relations, equivalence classes, boundary/edge cases). As illustrated in Figure \ref{fig:main_fig}, EvoSyn consists of three core stages: \textbf{(1) Deriving data filtering strategy}: deriving a reliable strategy using evolutionary algorithms. \textbf{(2) Data synthesis and filtering}: synthsizing data and filtering them with the derived strategy. \textbf{(3) Model training}: training models on the filtered synthetic data. The objective of EvoSyn is to establish a effective and automated mechanism that systematically enhances the reliability of synthetic data.

\subsection{Deriving data filtering strategy}

\begin{figure}[h]
\centering
\includegraphics[width=0.65\textwidth]{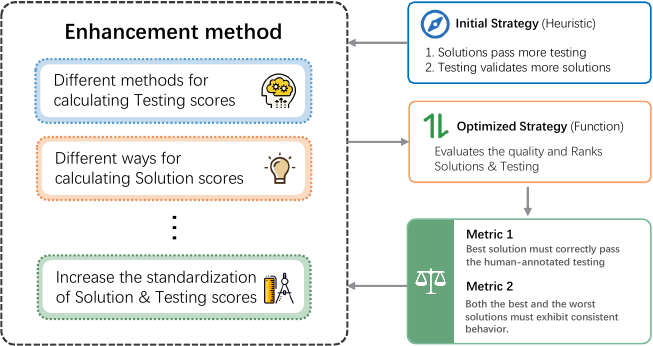}
\caption{Given an initial strategy, the evolutionary algorithm iteratively optimizes it across multiple iterations. Each newly generated strategy is evaluated against our two criteria to determine its effectiveness. The model autonomously explores diverse optimization approaches, ensuring a balance between exploration and exploitation throughout the process.}
\label{fig:evolution}
\vspace{-10pt}
\end{figure}

In the context of synthetic data, the central challenge is creating effective \textit{testing} mechanisms that can reliably verify candidate solutions. A high-quality data instance typically consists of two components: a problem description and its corresponding testing set. Producing reliable testings is highly difficult because testings must not only reflect an understanding of the problem but also need to steadily distinguish correct from incorrect solutions. If a testing cannot differentiate solution quality, the instance becomes unreliable even if the problem itself appears well-formed. Therefore, improving the reliability of testings is the central focus of our method.

We require filtered data to satisfy two conditions: (1) the problem must be solvable, and (2) the testing must reliably distinguish correct from incorrect solutions. In practice, the second condition is more challenging. Reliable testings must consistently and correctly distinguish between correct and wrong solutions, whereas proving that a problem is solvable only requires the existence of at least one solution that passes. Hence, reliability is the cornerstone of strategy design and optimization.
Formally, we model a data filtering strategy as a function that evaluates the quality of a set of testings. To derive such a strategy, we leverage seed data with human-annotated problems and testings. A relatively strong model is tasked with generating multiple candidate solutions for each problem, as well as additional testings based only on the problem description. These solutions and testings serve as inputs to the filtering strategy, which outputs a ranked list of solutions (optional) and testings.

Inspired by the success of \citet{novikov2025alphaevolve}, we adopt an evolutionary algorithm to iteratively improve the strategy. Evolutionary algorithms can balance exploration and exploitation. Following \citet{novikov2025alphaevolve} and \citet{openevolve}, our implementation combines the MAP-Elites algorithm~\citep{mouret2015illuminating} with island-based population models~\citep{romera2024mathematical, tanese1989distributed}, enabling optimization over user-defined feature dimensions while maintaining population diversity.
As in \citet{novikov2025alphaevolve}, the evolutionary process requires an initial filtering strategy, which \textit{need not be optimal}. We design the initial strategy according to two intuitive principles:
(1) solutions that pass more testings are considered better;
(2) testings that validate more solutions are considered better.
Although this initialization is imperfect, for example, a testing that passes all solutions is likely uninformative, it suffices to bootstrap the evolutionary process, which will refine and correct such limitations.
The evolutionary process also requires an \textit{evaluator}, whose role is to assess strategy quality with respect to user-defined criteria. We define a good strategy as one that ranks testings in close agreement with human-annotated testings on seed data across diverse candidate solutions. Specifically, the method for evaluating the quality of a strategy includes two strict criteria:

\text{~~~~~~• }\texttt{Criterion-1}: the top-ranked solution produced by the strategy must correctly pass the human-annotated testing in the seed data.

\text{~~~~~~• }\texttt{Criterion-2}: for the ranked solutions, both the best and the worst solutions must exhibit consistent behavior on the annotated testing and on the best testing selected by the strategy.

Figure \ref{fig:evolution} illustrates the actual workflow of our core method. Starting from an initial strategy, in each iteration the model explores various ways to obtain a better filtering strategy. From analyzing several relatively high-scoring strategies, we observe that the model explores multiple directions, such as refining the computation of solution quality and experimenting with different weighting schemes for testing. After each attempt, the evaluator assesses whether the new strategy satisfies our two predefined criteria on every instance in the seed data, and the proportion of satisfied cases is then used as the final score of the strategy, guiding the next round of evolution and refinement.

\begin{figure}[h]
\vspace{-10pt}
\centering
\includegraphics[width=0.6\textwidth]{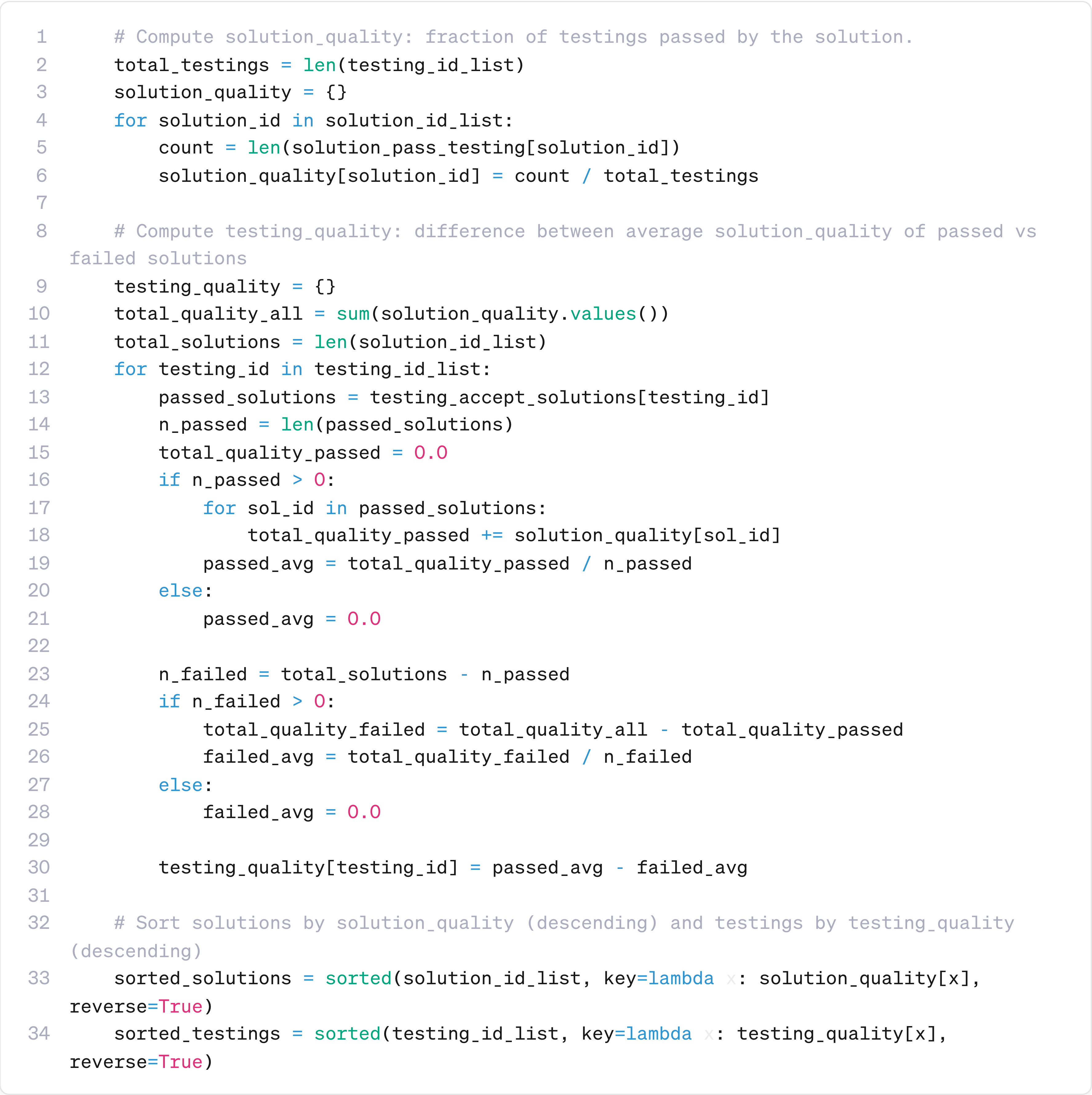}
\caption{The best strategy explored by model on LiveCodeBench.} 
\label{fig:best_program}
\vspace{-10pt}
\end{figure}

Remarkably, the evolutionary process yields multiple elegant and effective strategies. Figure \ref{fig:best_program} presents the best strategy evolved by model. The strategy scores each solution by the number of tests it passes, while testing scores are based on discriminative power (i.e., the gap between solutions' score that pass and fail), with both solution and testing scores normalized before computing discriminative power. Apart from this best one, model could explore various ways of computing testing scores. For example, \textit{TF-IDF-like approach}: solutions that pass difficult testings receive higher scores, where difficulty is defined as testings passed by only few solutions; \textit{Coverage-based approach}: solutions are rewarded simply for passing more testings, while testing quality is measured by its discriminative power (i.e., the score gap between solutions that pass and those that fail); \textit{Inverse filtering approach}: contrary to the initial strategy, testings that fewer solutions can pass are considered better; \textit{Exclusion-based approach}: the contribution of a testing is measured without its own influence, by weighting solutions that pass all other testings and \textit{Hardness-Aware approach}: solutions are ranked by test strictness and pass
count, penalizing all-or-none tests to select the strongest solution and most discriminative tests. The details of these strategies are illustrated in the Appendix \ref{subsec:optimized_strategies}.

These evolved strategies demonstrate strong internal logic and significantly improve upon manually designed baselines, showing that evolutionary search can efficiently discover high-quality filtering strategies.

\subsection{Data synthesis and filtering}

With a robust filtering strategy in place, we proceed to data synthesis and filtering. Specifically, we first synthesize new problems to replace the human-annotated seed data. To ensure the generated problems are compatible with the filtering strategy, we provide seed instances as in-context examples to guide problem generation.

After deduplication, the synthesized problems form a new set $D$. For each problem in $D$, we generate $M$ candidate solutions and $N$ candidate testings, which serve as inputs to the filtering strategy. The strategy ranks both solutions and testings. We then perform a final filtering step called \textit{Zero-Variance Pruning}: we discard instances in which the testings yield no ranking variation. Such cases typically indicate either unreliable testings or trivial problems where all testings perform equally well. In both scenarios, discarding the instance is justified.

\subsection{Model training}

Following the above steps, we obtain a reliable synthetic dataset containing problem descriptions and their associated testings. As a byproduct, we also retain the strongest solutions generated by the model. This dataset can be leveraged in various training paradigms, such as RLVR and model distillation, thereby boosting downstream model performance.

\section{Experiments}
\subsection{Experimental Setup}
This section presents a comprehensive set of experiments designed to validate the effectiveness of our proposed method. To address the specific challenge of verifiable problem synthesis, which is the core focus of our work, we conduct evaluations on two distinct and representative benchmarks: LiveCodeBench~\citep{jain2024livecodebench} and AgentBench-OS~\citep{liu2023agentbench}. LiveCodeBench is a highly challenging coding task benchmark, featuring a continuously updated collection of difficult programming problems. It has recently garnered significant attention within the large language model community due to its focus on real-time problem-solving capabilities. AgentBench-OS is a subset of the AgentBench benchmark, specifically designed to evaluate a model's performance in a realistic operating system environment. This benchmark assesses a model's ability to act as an intelligent agent and execute code to complete given tasks. The final performance is rigorously verified through a series of predefined tests.

To validate the effectiveness of our method, we conduct experiments on two representative training paradigms: reinforcement learning with verifiable rewards (RLVR)~\citep{guo2025deepseek} and model distillation. Considering both cost and task complexity, we employ DeepSeek-V3\footnote{Due to the excessively long chains-of-thought (CoT) produced by DeepSeek-R1 on algorithmic problems, which lead to slow inference, we use DeepSeek-V3 as the teacher model for the LiveCodeBench task. We also observe poor performance in instruction following during question generation.} and DeepSeek-R1 as teacher models for the two tasks, respectively. The teacher model is responsible for the entire data synthesis pipeline, including the filtering strategy and data generation.
During the evolutionary process for each task, we synthesize $M=16$ solutions and their corresponding testing for every problem instance. The maximum number of evolutionary iterations is set to 20.
\begin{figure}[h]
   \centering
   \makebox[\textwidth][c]{
      \subfloat[Evolutionary process over 20 iterations.]{\includegraphics[width=0.494\textwidth]{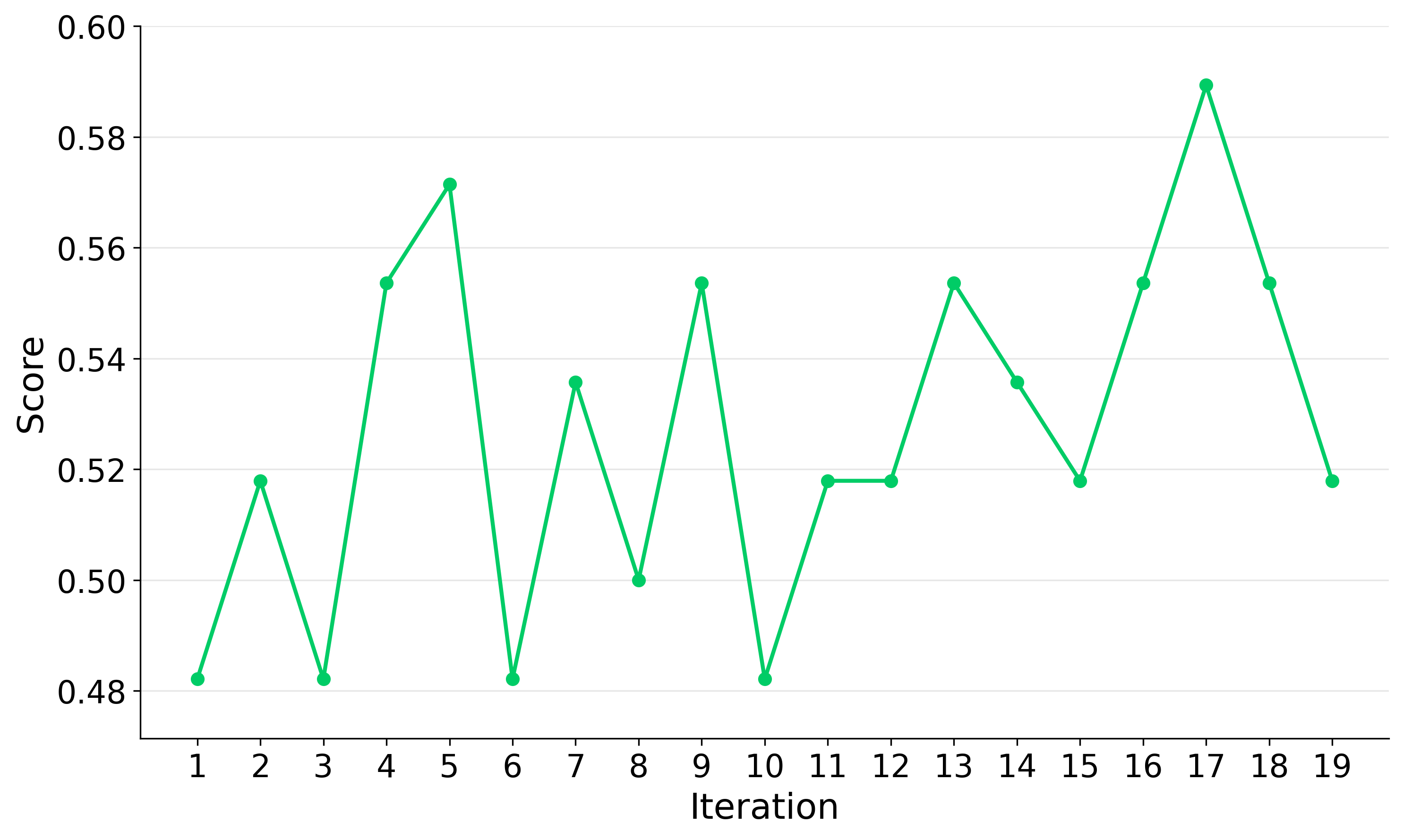}\label{fig:evolve_process}}
      \hspace{0.03\textwidth}
      \subfloat[Effect of different $(M,N)$.]{\includegraphics[width=0.34048\textwidth]{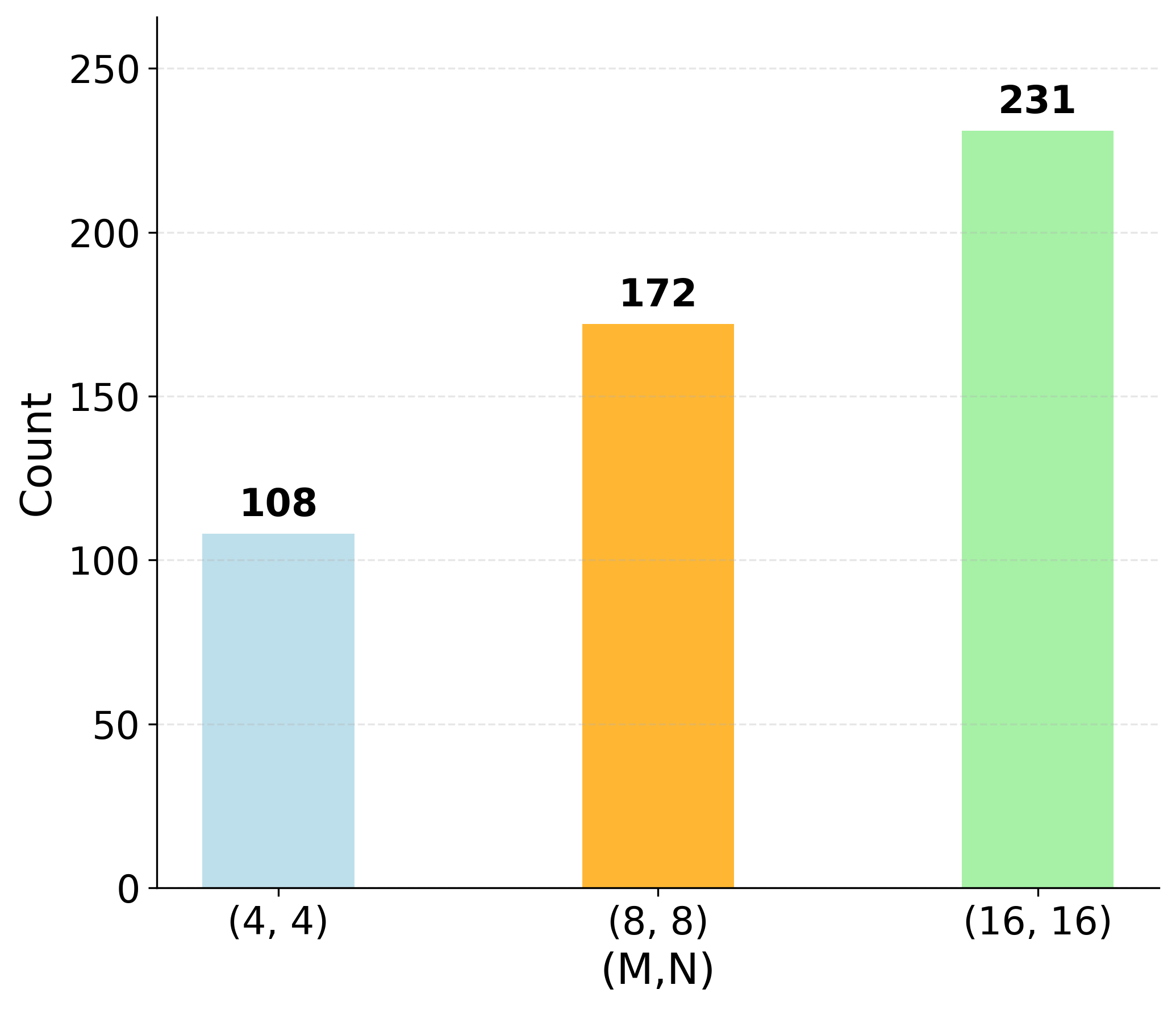}\label{fig:count_distribution}}
   }
   \caption{Evolutionary process and data-retention trade-off. (a) The evolutionary process consistently discovers stronger strategies, with the best strategy surpassing the initialization by over 10 percentage points within 20 iterations. Score denotes the ratio of seed data instances for which consistency verification is satisfied. (b) Increasing the number of $M$ and $N$ yields more usable, verifiable instances but incurs $O(MN)$ testing execution cost.}
   \label{fig:evolution_analysis}
   \vspace{-10pt}
\end{figure}

\subsection{Evolutionary Process}

In this set of experiments, we use the LivecodeBench task as an example to illustrate the effectiveness of our core evolutionary method in the data synthesis process. As shown in Figure~\ref{fig:evolve_process}, within the limit of 20 evolutionary iterations, and after excluding a few strategies that contained bugs, we frequently observe strategies outperforming the initial baseline. In particular, the best strategy exceeds the initial one by more than 10 percentage points, demonstrating both the model's ability to explore diverse strategies and the effectiveness of applying evolutionary algorithms to this problem.
Moreover, the overall trend of the evolutionary process shows a steady upward trajectory. This suggests that, with more iterations, there is a strong potential to discover even better filtering strategies to guide data synthesis, highlighting the feasibility of our approach.

\paragraph{Ablation study 1: Impact of $M$ and $N$} However, better strategies also imply stricter filtering standards. To investigate this, we apply the best evolved strategy to data synthesis while varying the value of $M$ and $N$. The choice of $M$ and $N$ has a significant impact on synthesis cost, since our method requires generating $M$ solutions and $N$ testings, followed by $M*N$ executions. This quadratic growth in cost makes it crucial to understand the relationship between $(M,N)$ and the amount of usable data ultimately obtained.
As shown in Figure~\ref{fig:count_distribution}, when applying to the same set of 1,250 problems, using $M=N=4$, $M=N=8$, and $M=N=16$ produces markedly different amounts of usable data. The reason is straightforward: with fewer samples, the likelihood of obtaining diverse solutions and testings decreases, making it harder to generate varied feedback and, consequently, to verify reliability.

\begin{wraptable}{r}{0.5\textwidth}
   \vspace{-12pt}
   \centering
   \small
   \caption{Consistency validation on $M\!=\!16$ solutions. We vary $K$ and validate the same strategy using the top-$K$ and bottom-$K$ solution subsets. Adding \texttt{Criterion-1} at $K\!=\!1$ yields the strictest check while requiring $8\times$ fewer executions ($\texttt{\#Exec}\!=\!4$ vs. 32) than omitting it at $K\!=\!8$. Increasing $K$ alone shows diminishing returns.}
   \label{tab:consistency_validation_results}
   \resizebox{!}{1.8cm}{
      \begin{tabular}{lccc}
      \toprule
      $K$ & \texttt{Criterion-1} & Score & \texttt{\#Exec} \\
      \midrule
      1 & No & 0.589 & 4 \\
      2 & No & 0.554 & 8 \\
      4 & No & 0.536 & 16 \\
      8 & No & 0.536 & 32 \\
      \hline
      \multicolumn{4}{c}{\cellcolor{gray!11}\textbf{\textit{Ours}}} \\
      1 & Yes & 0.482 & 4 \\
      8 & Yes & 0.482 & 32 \\
      \bottomrule
      \end{tabular}
   }
   \vspace{-8pt}
\end{wraptable}

\paragraph{Ablation Study 2: Is Consistency Validation Sufficient with Only the Best and Worst Solutions?}
Recalling our two evaluation criteria for strategy assessment: \texttt{Criterion-1}, the best solution must be correct; \texttt{Criterion-2} the performance of the best and worst solutions must agree on both the human-annotated test set and the strategy-selected best test. A natural question arises: are these criteria sufficient? To investigate, we vary the number of solutions used for evaluating a same strategy (we use the initial strategy as an example) and choosing $M=16$, considering the best and worst $K$ solutions with $K=1,2,4$ and $K=8$ (i.e., $M/2$). As shown in Table~\ref{tab:consistency_validation_results}, increasing $K$ indeed strengthens the evaluation criteria, reflects in lower overall scores. However, two observations emerge. First, the stricter constraint does not scale linearly with $K$: validating more solutions does not necessarily yield proportionally more accurate evaluations, largely due to randomness in solution sampling. Second, our setting combining the two criteria with $K=1$, is in fact stricter than the $K=8$ case, achieving both higher accuracy and significantly greater efficiency.

In our experiments, although the proposed method is in principle capable of generating unlimited data and producing highly reliable testings, from Figure~\ref{fig:count_distribution}, we observe a log-linear relationship between the number of usable data instances and the number of testing executions. The underlying reason lies in the difficulty of controlling the diversity of model outputs. Low diversity inevitably requires larger values of $M$ and $N$, which substantially increases the cost. In future work, we aim to further investigate methods to enhance output diversity while reducing synthesis costs. In addition, practical bottlenecks such as slow model inference, time-intensive unit test verification, and costly environment setup further constrain the scalability of our data synthesis. We therefore adopt $N=16$, yielding over 200 instances for LiveCodeBench and over 600 instances for AgentBench-OS. Despite this relatively small scale, training on these data still leads to substantial performance improvements.

\subsection{EvoSyn for RLVR}
This set of experiments demonstrates that synthetic data generated with our method can effectively improve model performance in the RLVR task. We construct three data settings based on 51 seed instances:

\text{~~~~~~• }$D^{\mathrm{EvoSyn}}$: Data filtered using our proposed data filtering strategy.

\text{~~~~~~• }$D^{\mathrm{random}}$: Data with exactly the same problems as $D^{\mathrm{EvoSyn}}$, but instead of using the filtering strategy, we randomly select one testing from the $N$ candidates as the final testing.

\text{~~~~~~• }$D^{\mathrm{EvoSyn^{relaxed}}}$: Data obtained by relaxing the \textit{Zero-Variance Pruning}, to investigate the necessity of our method's final filtering condition, which excludes instances that have not undergone ranking.

In particular, we analyze the number of unit tests per synthesized data in $D^{\mathrm{EvoSyn}}$. As shown in Figure~\ref{fig:unit_test_num}, the synthesized data contain an average of 11.5 unit tests, including various edge cases such as extremely long inputs. To mitigate the strong dependency on long-context capability imposed by such edge cases, we further adjust the testing generation process: instead of asking the model to directly produce unit tests, we require it to output code from which unit tests can be constructed. This not only preserves the diversity of unit tests but also ensures that the number of tests per problem remains sufficiently large.

\begin{table}[h]
   \caption{RLVR results on LiveCodeBench: Training on EvoSyn-filtered data ($D^{\mathrm{EvoSyn}}$) consistently improves accuracy across models, outperforming random selection ($D^{\mathrm{random}}$) and the relaxed variant ($D^{\mathrm{EvoSyn^{relaxed}}}$). $\Delta$ denotes absolute gain over the baselines.}
   \label{tab:rlvr_results}
   \begin{center}
   \vspace{-5pt}
   \resizebox{!}{2.8cm}{
      \begin{tabular}{l c S[table-format=3] S[table-format=2.1] >{\color{lightgreen}}S[table-format=+2.1]}
      \toprule
      Model & Data Setting & {Dataset Size} & {Accuracy} & \multicolumn{1}{c}{$\Delta$}\\
      \midrule
      \multicolumn{5}{c}{\cellcolor{gray!11}\textbf{\textit{Baseline}}} \\
      DeepSeek-V3& - & \multicolumn{1}{c}{-} & 36.3 & \multicolumn{1}{c}{-} \\
      Qwen3-4B & - & \multicolumn{1}{c}{-} & 17.0 & \multicolumn{1}{c}{-} \\
      Llama-3.1-8B & \multicolumn{1}{c}{-} & \multicolumn{1}{c}{-} & 1.6 & \multicolumn{1}{c}{-} \\
      Qwen3-8B & \multicolumn{1}{c}{-} & \multicolumn{1}{c}{-} & 16.5 & \multicolumn{1}{c}{-} \\
      \midrule
      \multicolumn{5}{c}{\cellcolor{gray!11}\textbf{\textit{RLVR Models}}} \\
      Qwen3-4B & \makebox[\Dcolw][l]{\hspace{-12pt}$D^{\mathrm{EvoSyn}}$} & 231 & 22.0 & +5.0 \\
      Qwen3-4B & \makebox[\Dcolw][l]{\hspace{-12pt}$D^{\mathrm{random}}$} & 231 & 19.9 & +2.9 \\
      Llama-3.1-8B & \makebox[\Dcolw][l]{\hspace{-12pt}$D^{\mathrm{EvoSyn}}$} & 231 & 15.7 & +14.1 \\
      Llama-3.1-8B & \makebox[\Dcolw][l]{\hspace{-12pt}$D^{\mathrm{random}}$} & 231 & 11.1 & +9.5 \\
      Qwen3-8B & \makebox[\Dcolw][l]{\hspace{-12pt}$D^{\mathrm{EvoSyn}}$} & 231 & \textbf{24.8} & +8.3 \\
      Qwen3-8B & \makebox[\Dcolw][l]{\hspace{-12pt}$D^{\mathrm{random}}$} & 231 & 21.1 & +4.6 \\
      Qwen3-8B & \makebox[\Dcolw][l]{\hspace{-12pt}$D^{\mathrm{EvoSyn^{relaxed}}}$} & 256 & 24.4 & +7.9 \\
      \bottomrule
      \end{tabular}
   }
   \vspace{-10pt}
   \end{center}
\end{table}
\paragraph{Results}
We conduct reinforcement learning experiments on Qwen3-8B, Qwen3-4B, and Llama-3.1-8B using GRPO. 
As shown in Table \ref{tab:rlvr_results}, training on our synthesized dataset $D^{\mathrm{EvoSyn}}$ consistently yields significant performance gains across all models. Notably, on Llama-3.1-8B, we observe a substantial improvement of 14.1\%. This demonstrates the effectiveness of our method in synthesizing reliable data. 

\begin{figure}[h]
\vspace{-5pt}
\centering
\subfloat[Llama-3.1-8B]{\includegraphics[width=0.33\textwidth]{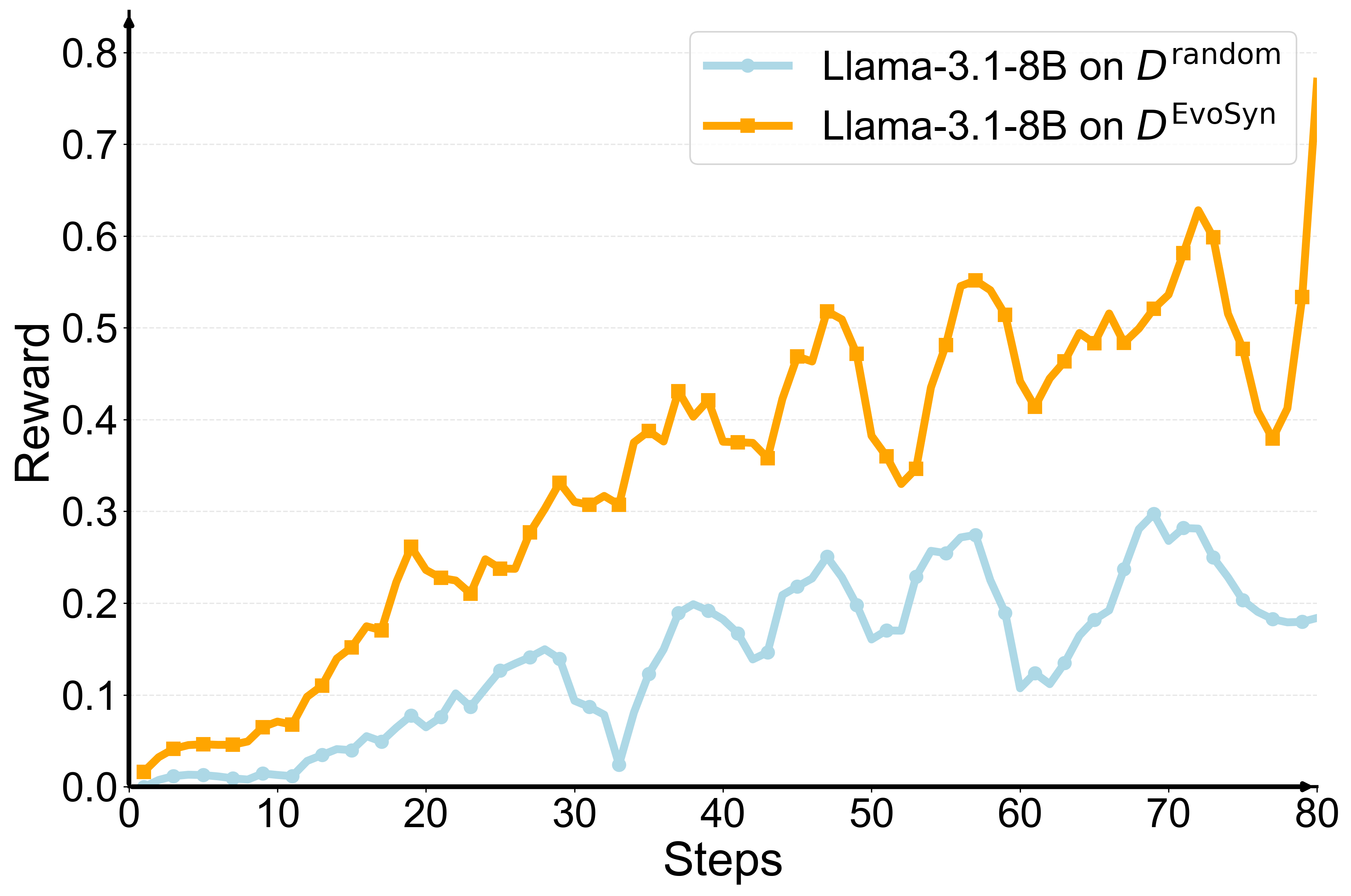}\label{fig:llama-3-1-8b}}
\hfill
\subfloat[Qwen3-8B]{\includegraphics[width=0.33\textwidth]{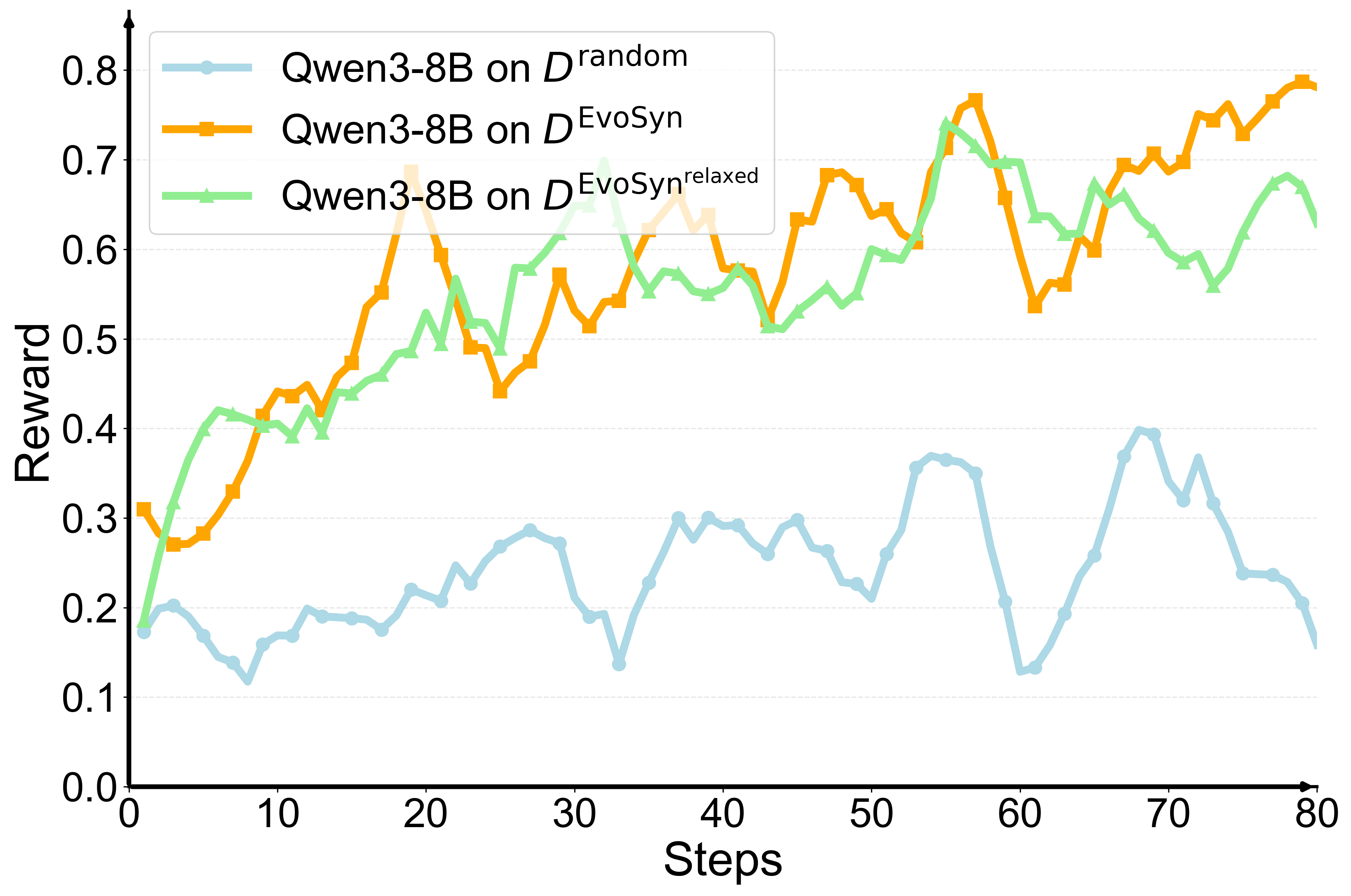}\label{fig:qwen3-8b}}
\hfill
\subfloat[Qwen3-4B]{\includegraphics[width=0.33\textwidth]{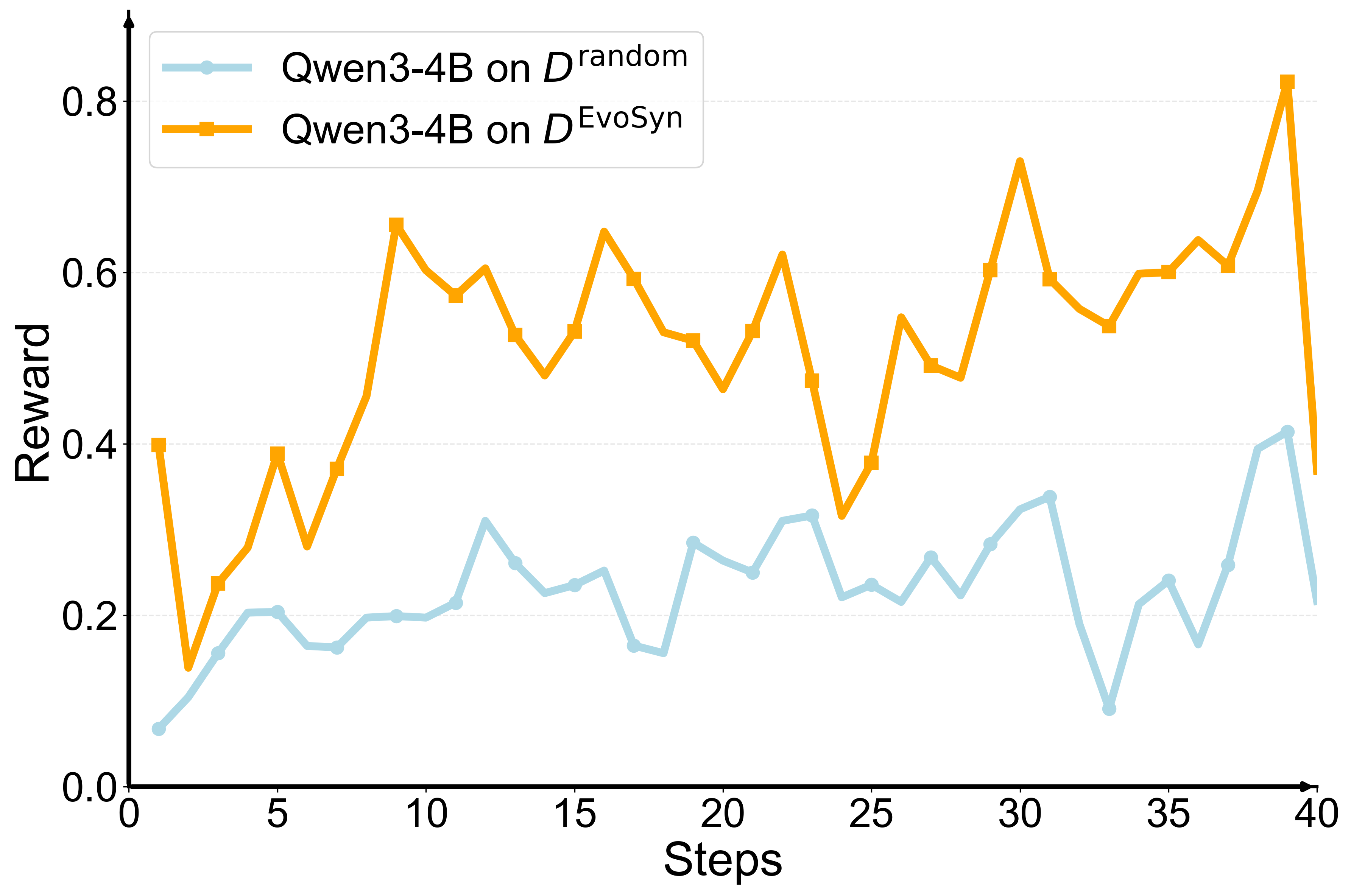}\label{fig:reward_comparison}}
\caption{RLVR reward curves comparison across models. EvoSyn-filtered data ($D^{\mathrm{EvoSyn}}$) yields faster, steadier reward growth than random selection ($D^{\mathrm{random}}$).}
\label{fig:evolution_analysis}
\vspace{-10pt}
\end{figure}

\paragraph{Ablation Study 3: What drives this advantage?} 
To further demonstrate the effectiveness of our method, we compare it against randomly synthesized data without filtering. As shown in Table~\ref{tab:rlvr_results}, although training with randomly synthesized data on Qwen3-8B also yields some improvement, indicating that a portion of the data is indeed learnable, the performance still lags significantly behind that achieved with our filtered dataset.
This result can be further analyzed through the reward dynamics during training. As illustrated in Figure~\ref{fig:evolution_analysis}, training on $D^{\mathrm{EvoSyn}}$ exhibits a steady and meaningful increase in reward, whereas training on $D^{\mathrm{random}}$ struggles to achieve consistent reward growth. This comparison highlights that the data constructed by our method is substantially more learnable for the model.

\paragraph{Ablation Study 4: Is the \textit{Zero-Variance Pruning} necessary?} 
In addition, we analyze the differences between $D^{\mathrm{EvoSyn}}$ and $D^{\mathrm{EvoSyn^{relaxed}}}$. By design, $D^{\mathrm{EvoSyn}}$ is a strict subset of $D^{\mathrm{EvoSyn^{relaxed}}}$. We manually examine the 25 additional instances present to $D^{\mathrm{EvoSyn^{relaxed}}}$ but not in $D^{\mathrm{EvoSyn}}$, and find that nearly all of them were overly simple problems. On average, their solution code lengths are only a dozen lines, and in some cases, the $M$ solutions sampled at temperature 1.0 are completely identical. 

\begin{wraptable}{r}{0.5\linewidth}
   \vspace{-12pt}
   \caption{Model distillation results on AgentBench-OS: EvoSyn-filtered data ($D^{\mathrm{EvoSyn}}$) yields large gains across students, outperforming random selection ($D^{\mathrm{random}}$). Remarkably, all students exceed the teacher (DeepSeek-R1, 30.1).}
   \label{tab:distillation_results}
   \centering
   \resizebox{!}{2.6cm}{
      \begin{tabular}{lc S[table-format=+2.1]  >{\color{lightgreen}}S[table-format=+2.1]}
      \toprule
      Model & Data Setting & Accuracy & \multicolumn{1}{c}{$\Delta$}\\
      \midrule
      \multicolumn{4}{c}{\cellcolor{gray!11}\textbf{\textit{Baseline}}} \\
      DeepSeek-R1 & - & 30.1 & \multicolumn{1}{c}{-} \\
      Qwen3-4B & - & 1.0 & \multicolumn{1}{c}{-} \\
      Llama-3.1-8B & - & 1.0 & \multicolumn{1}{c}{-} \\
      Qwen3-8B & - & 1.0 & \multicolumn{1}{c}{-} \\
      \midrule
      \multicolumn{4}{c}{\cellcolor{gray!11}\textbf{\textit{Distilled Models}}} \\
      Qwen3-4B & \makebox[\Dcolw][l]{\hspace{-12pt}$D^{\mathrm{EvoSyn}}$} & 40.0 & +39.0\\
      Qwen3-4B & \makebox[\Dcolw][l]{\hspace{-12pt}$D^{\mathrm{random}}$} & 36.0 & +35.0 \\
      Llama-3.1-8B & \makebox[\Dcolw][l]{\hspace{-12pt}$D^{\mathrm{EvoSyn}}$} & 37.6 & +36.6 \\
      Llama-3.1-8B & \makebox[\Dcolw][l]{\hspace{-12pt}$D^{\mathrm{random}}$} & 22.0 & +21.0 \\
      Qwen3-8B & \makebox[\Dcolw][l]{\hspace{-12pt}$D^{\mathrm{EvoSyn}}$} & \textbf{44.9}\hspace{-8pt} & +43.9 \\
      Qwen3-8B & \makebox[\Dcolw][l]{\hspace{-12pt}$D^{\mathrm{random}}$} & 32.8 & +31.8 \\
      \bottomrule
      \end{tabular}
   }
   \vspace{-30pt}
\end{wraptable}

This observation validates the rationale behind the \textit{Zero-Variance Pruning} in our method, which removes overly simple problems that provide little value for model learning. Such trivial problems are particularly problematic for the RLVR paradigm, as they prevent proper computation of the advantage. Selected example is provided in the Appendix \ref{subsec:trivial_problem}.

\subsection{EvoSyn for Model Distillation}

Model distillation has been widely adopted in the field due to its effectiveness and high efficiency, making it a powerful alternative to reinforcement learning, especially when the latter's training costs become prohibitive. Similar to RLVR, this method critically depends on a high-quality set of problems and reliable testings. This robust evaluation mechanism is essential for accurately filtering the correct responses from a teacher model.
In this experiment, we select the AgentBench-OS task, which is a highly realistic agent task requiring multi-turn, complex reasoning. These abilities are often a significant weakness for many models, particularly smaller ones. Due to the task's complex environment setup (the need for isolated Docker environments), the associated costs are prohibitively high, making RLVR-based training difficult. Therefore, we experimentally validate the effectiveness of our proposed method within a model distillation pipeline. We use OpenHands~\citep{wang2024openhands} as the agent framework for our models. We filter the original AgentBench-OS dataset due to the presence of samples with stringent time requirements or permission issues. From the initial 144 data points, we retain 129, which are subsequently used as both the evaluation set and the seed data for our proposed method. This curation process ensures the reliability and reproducibility of our experimental results by focusing on a stable and accessible subset of the benchmark.

\paragraph{Results}

Based on our method, we synthesize 673 data instances and obtain the corresponding outputs from DeepSeek-R1. Using this synthetic dataset, we train Qwen3-4B, Llama-3.1-8B, and Qwen3-8B.
As shown in Table~\ref{tab:distillation_results}, all models exhibit substantial performance improvements after training. This not only highlights the weaker baseline performance of smaller models on complex, multi-turn, long-chain reasoning tasks but also clearly demonstrates the effectiveness of our synthetic data generation method. Furthermore, training on data synthesized by our method significantly outperforms training on randomly synthesized data, indicating that our approach is more effective at filtering usable data in complex, real-world agent tasks.

\section{Conclusion}
We introduce EvoSyn, a task-agnostic evolutionary data synthesis framework that focuses on synthesizing verifiable data for executably-checkable tasks by evolving robust filtering strategies from minimal seed supervision via a consistency-based evaluator. By turning ad hoc filtering into principled strategy optimization, EvoSyn assembles coherent, verifiable training instances that transfer across domains. On LiveCodeBench (RLVR) and AgentBench-OS (distillation), training on EvoSyn-filtered data yields substantial gains and superior learning dynamics across Llama-3.1-8B and Qwen3-4B/8B, with distilled students surpassing the teacher on complex multi-turn agentic tasks. Ablations confirm the value of strategy evolution and Zero-Variance Pruning, and characterize cost–quality trade-offs in $M*N$ execution. Limitations include verification/execution cost and output-diversity bottlenecks. Future work will scale population search, improve diversity-aware generation, and broaden verification tooling and domains.

\bibliography{iclr2026_conference}
\bibliographystyle{iclr2026_conference}

\clearpage
\appendix
\section{Appendix}

\subsection{Prompts}
This subsection provides the exact prompts used in our synthesis pipeline. We include the LiveCodeBench testing prompt and the AgentBench‑OS prompts (There are two different types of responses, and we provide separate prompts for each type). They can be used to reproduce our data generation and to inspect task‑specific constraints and formatting requirements.
\begin{figure}[H]
   \vspace{-15pt}
   \centering
   \includegraphics[width=0.8\textwidth]{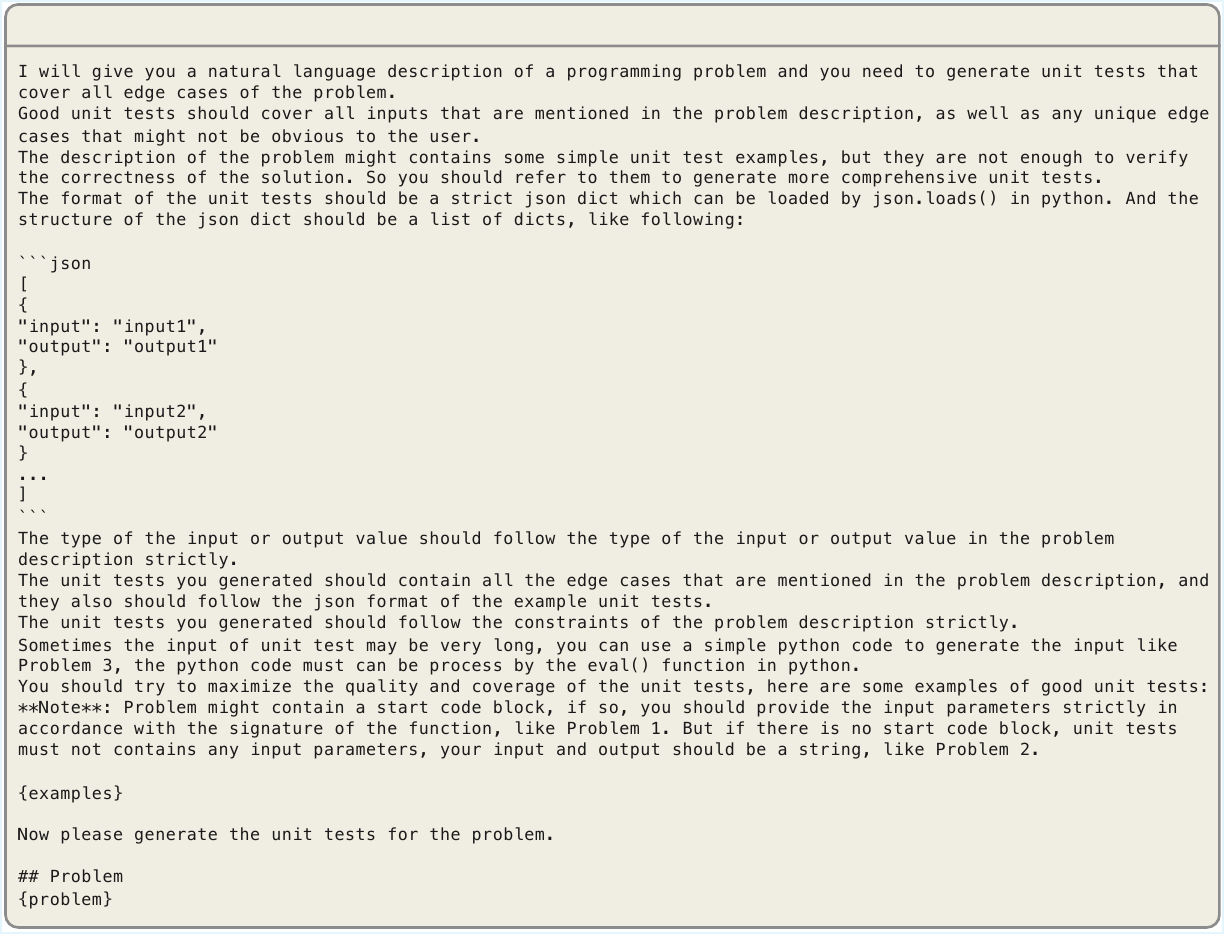}
   \caption{Prompt for testing generation on LiveCodeBench.}
   \label{fig:livecode_testing_prompt}
   \vspace{-15pt}
\end{figure}

\begin{figure}[H]
   \centering
   \includegraphics[width=0.6\textwidth]{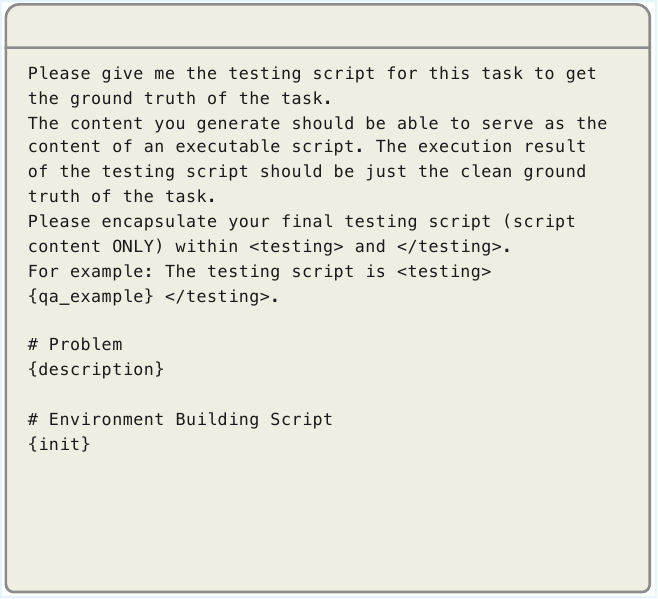}
   \caption{Prompt for testing generation on AgentBench-OS QA task.}
   \label{fig:agentbench_qa_prompt}
\end{figure}

\begin{figure}[H]
   \centering
   \includegraphics[width=0.6\textwidth]{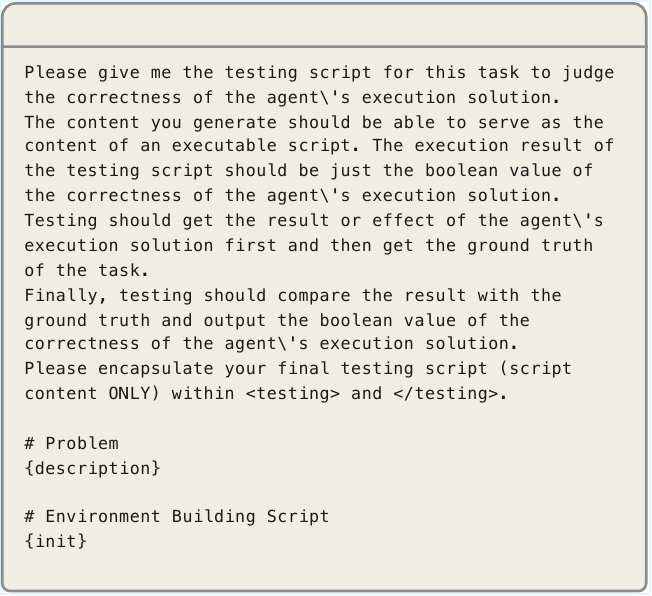}
   \caption{Prompt for testing generation on AgentBench-OS Execution task.}
   \label{fig:agentbench_exec_prompt}
\end{figure}

\subsection{Unit-test count distribution}
\label{subsec:unit_test_count_distribution}
We analyze the number of unit tests attached to each synthesized problem to characterize the strength and granularity of our automated evaluation. Counts include both standard checks and long‑input edge‑case tests. The distribution is broad (mean 11.5 per problem), indicating heterogeneous coverage and difficulty, which helps produce more stable and discriminative reward signals for ranking solutions and selecting tests.
\begin{figure}[H]
   \centering
   \includegraphics[width=0.7\textwidth]{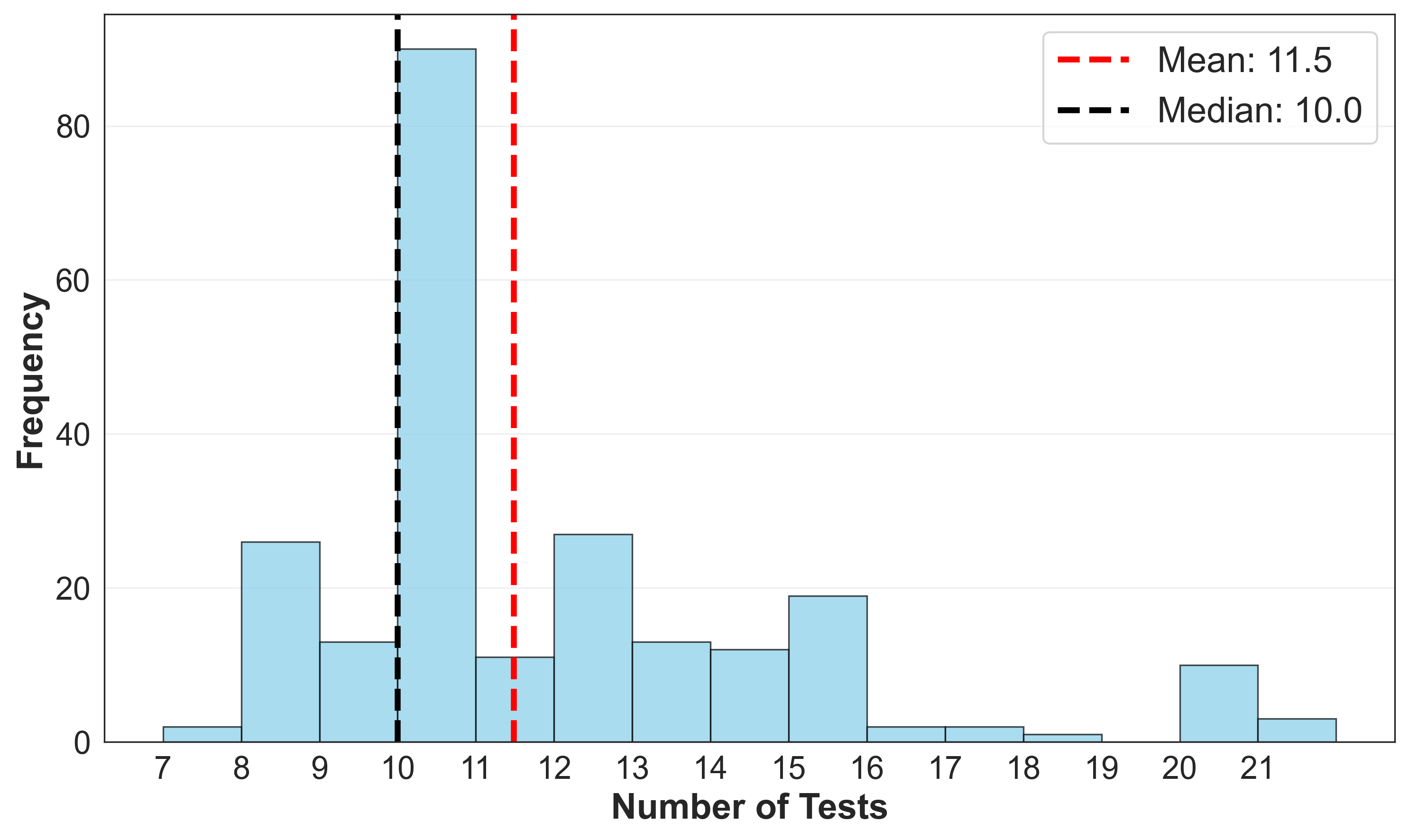}
   \caption{Unit-test count distribution per synthesized problem (mean \(11.5\)); includes long-input edge cases.}
   \label{fig:unit_test_num}
\end{figure}

\subsection{Optimized strategies}
\label{subsec:optimized_strategies}
We outline several evolved scoring strategies that complement the best program in Figure~\ref{fig:best_program}: TF‑IDF‑like weighting, coverage‑based scoring, inverse filtering, exclusion‑based attribution, and hardness‑aware ranking. Each aims to improve discriminative power and solution‑ranking consistency on seed data.

\begin{figure}[H]
   \centering
   \includegraphics[width=0.8\textwidth]{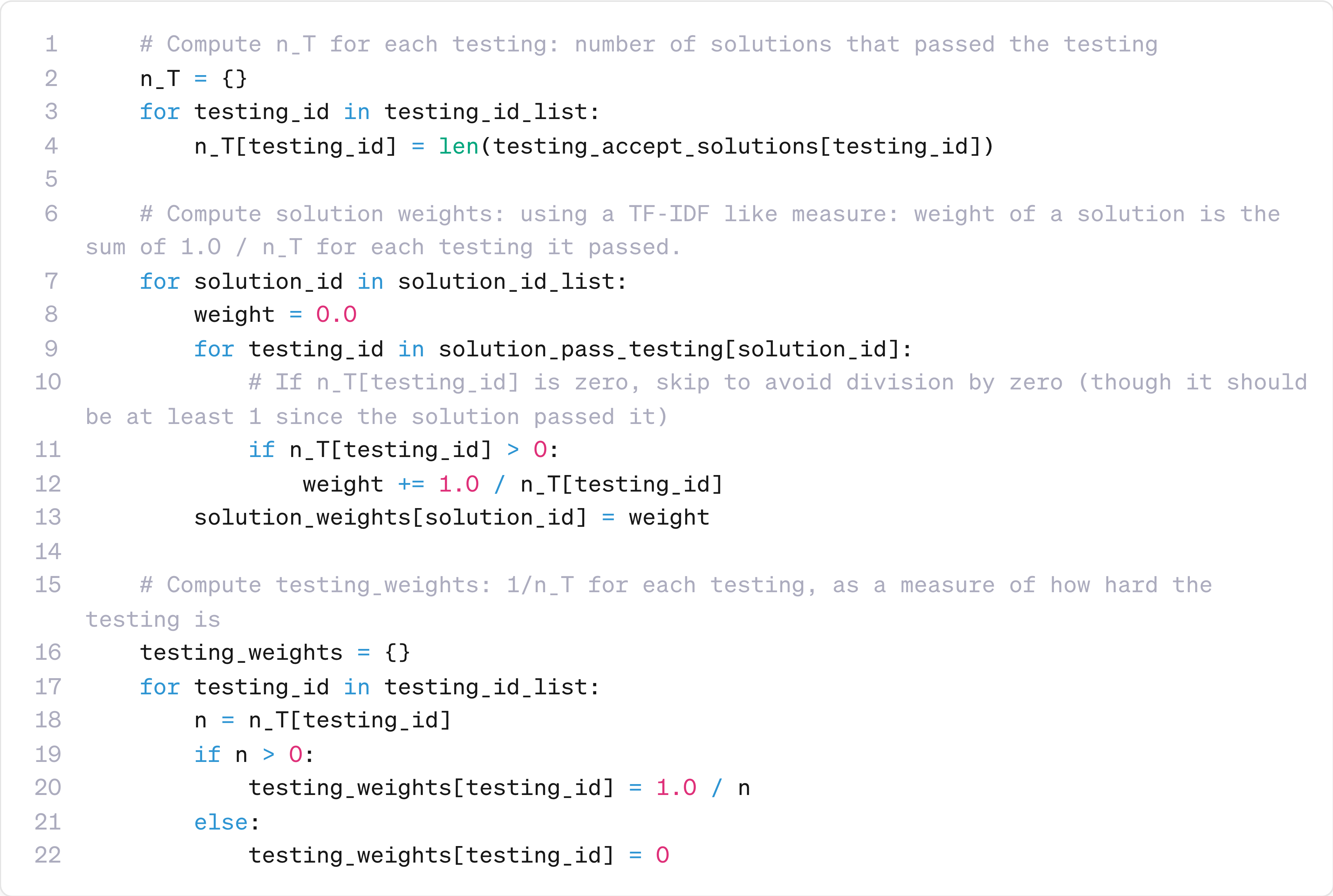}
   \caption{TF-IDF-like approach. Solutions that pass difficult testings receive higher scores, where difficulty is defined as testings passed by only a few solutions.}
   \label{fig:tfidf_like}
\end{figure}

\begin{figure}[H]
   \vspace{-20pt}
   \centering
   \includegraphics[width=0.8\textwidth]{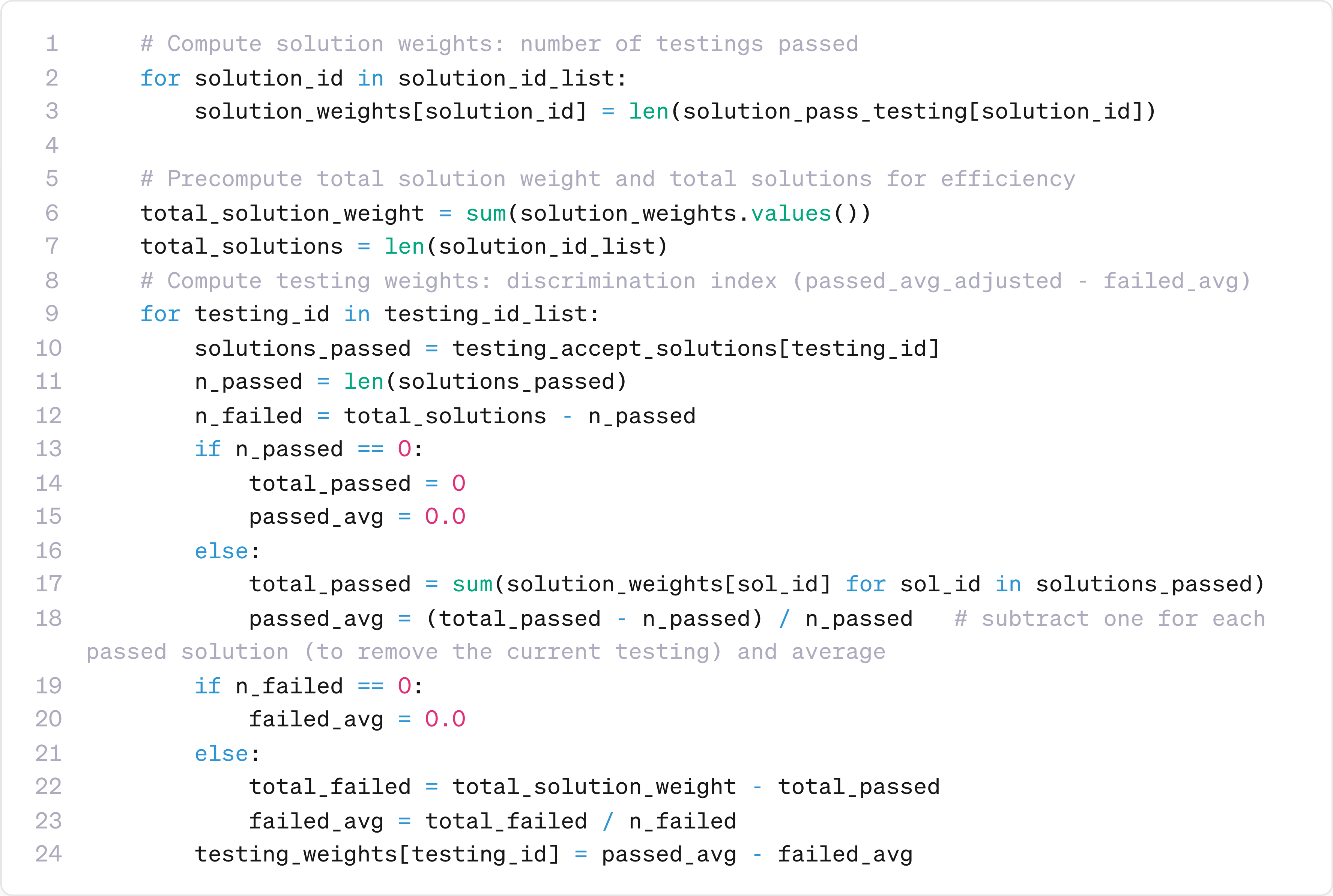}
   \caption{Coverage-based approach. Solutions are rewarded simply for passing more testings, while testing quality is measured by its discriminative power (i.e., the score gap between solutions that pass and those that fail).}
   \label{fig:coverage_based}
   \vspace{-20pt}
\end{figure}

\begin{figure}[H]
   \vspace{-20pt}
   \centering
   \includegraphics[width=0.8\textwidth]{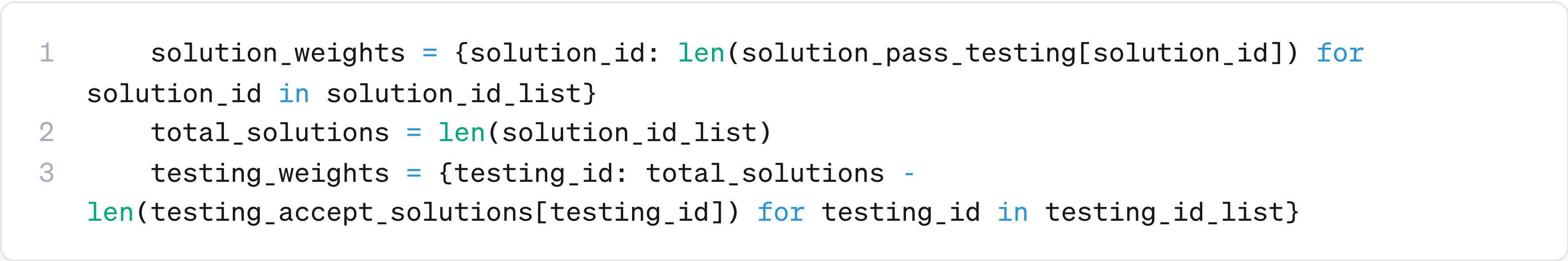}
   \caption{Inverse filtering approach. Contrary to the initial strategy, testings that fewer solutions can pass are considered better.}
   \label{fig:inverse_filtering}
\end{figure}

\begin{figure}[H]
   \centering
   \includegraphics[width=0.8\textwidth]{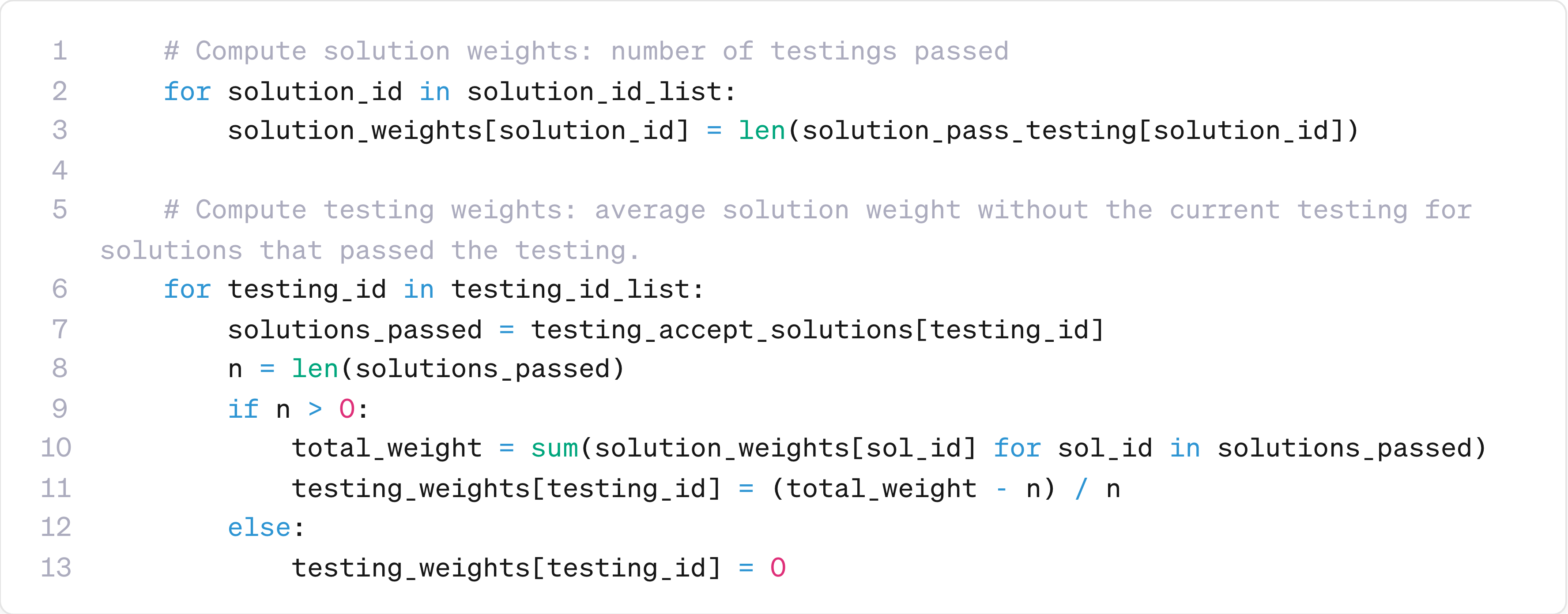}
   \caption{Exclusion-based approach. The contribution of a testing is measured without its own influence, by weighting solutions that pass all other testings.}
   \label{fig:exclusion_based}
\end{figure}

\begin{figure}[H]
   \vspace{-10pt}
   \centering
   \includegraphics[width=0.8\textwidth]{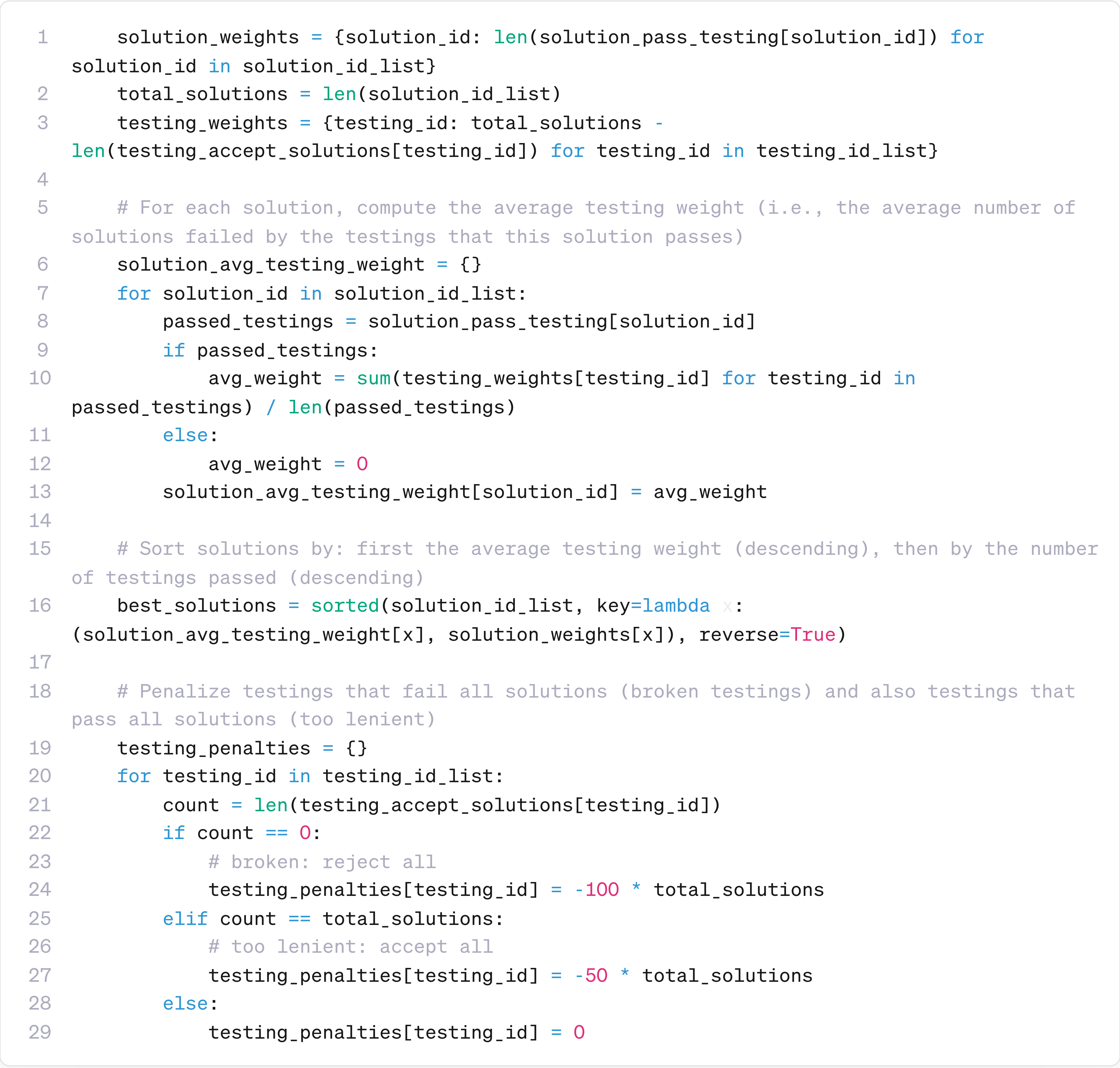}
   \caption{Hardness-Aware approach. Solutions are ranked by test strictness and pass count, penalizing all-or-none tests to select the strongest solution and most discriminative tests.}
   \label{fig:standardized_discriminative}
\end{figure}

\subsection{Trivial problem}
\label{subsec:trivial_problem}
We manually examine the 25 additional instances present to $D^{\mathrm{EvoSyn^{relaxed}}}$ but not in $D^{\mathrm{EvoSyn}}$, and find that nearly all of them were overly simple problems. Here we provide one of the examples, the quesiton is just a simple determination of whether some numbers are all even or not.

\begin{figure}[H]
   \centering
   \includegraphics[width=0.9\textwidth]{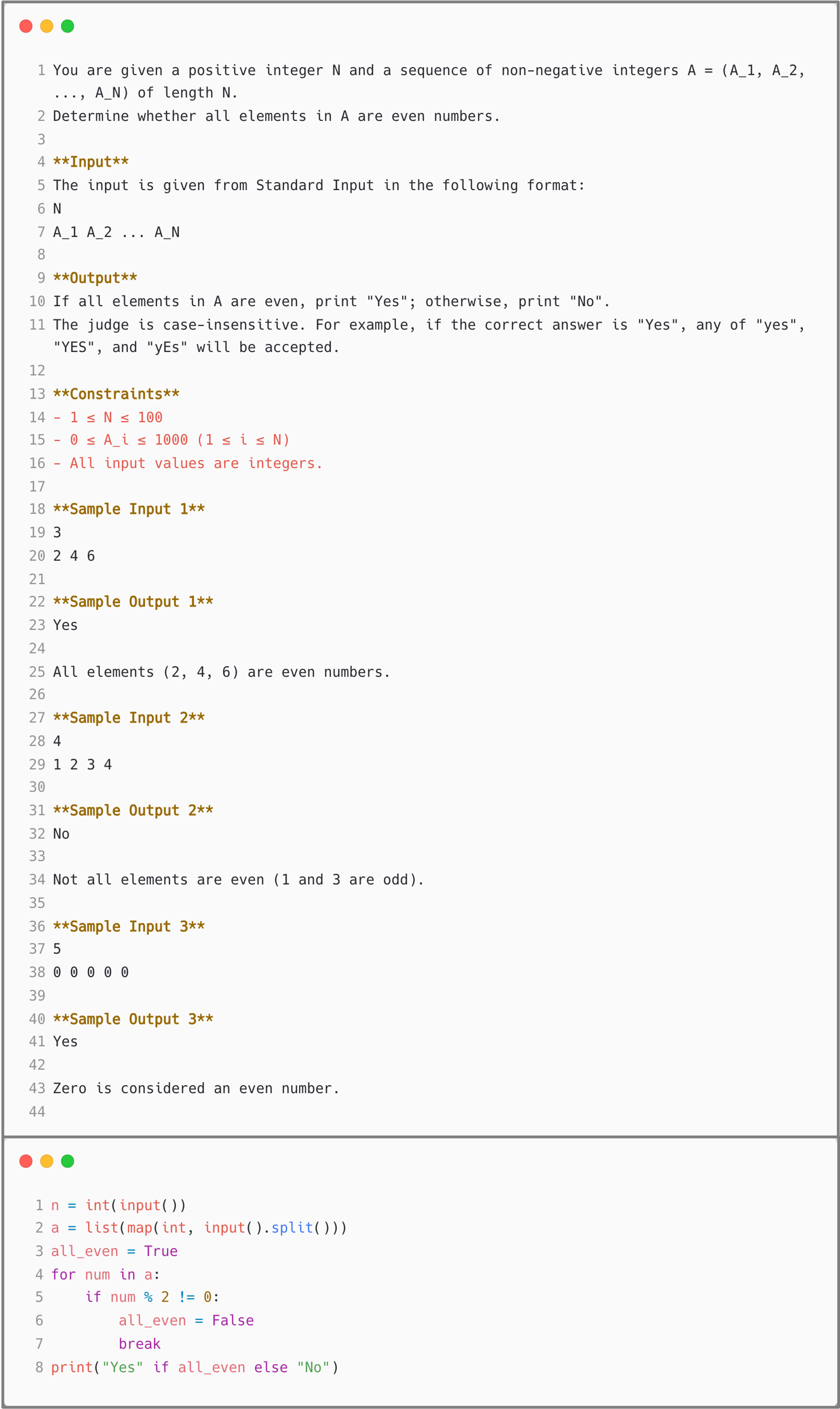}
   \caption{An example of a trivial problem present in $D^{\mathrm{EvoSyn^{relaxed}}}$ but not in $D^{\mathrm{EvoSyn}}$, containing quesiton description and solution code.}
   \label{fig:trivial_problem}
\end{figure}

\subsection{AI Usage Statement}
AI tools were used solely to assist with writing and polishing the main manuscript text. All core research content—including the ideas, problem formulation, methodology and algorithm design, data synthesis framework, experimental design and execution, implementation, evaluation, and analysis—was conceived, conducted, and validated exclusively by the authors. No AI systems were involved in generating ideas, designing or running experiments, or producing any core research results.

\end{document}